\pdfoutput=1

\documentclass[11pt]{article}

\usepackage[]{ACL2023}

\usepackage{times}
\usepackage{latexsym}
\usepackage{multirow}
\usepackage{url}
\usepackage{latexsym}
\usepackage{balance}
\usepackage{bm}
\usepackage{amssymb}
\usepackage{amsmath}
\usepackage{url}
\usepackage{makecell}
\usepackage{multirow}
\usepackage{subfig}
\usepackage{graphicx} 
\usepackage{color}
\usepackage{colortbl}
\usepackage{textcomp}
\usepackage{CJK}
\usepackage{enumitem}
\usepackage{booktabs}
\usepackage{microtype}
\usepackage{arydshln}
\usepackage{amsfonts}
\usepackage{eufrak}
\usepackage[T1]{fontenc}

\usepackage[utf8]{inputenc}

\usepackage{microtype}
\usepackage{pifont}
\usepackage{inconsolata}

\usepackage{graphicx}
\def\emojihappy{\raisebox{-0.55ex}{\includegraphics[width=1.3em]{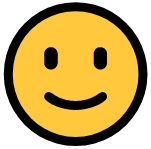}}}
\def\emojisad{\raisebox{-0.55ex}{\includegraphics[width=1.3em]{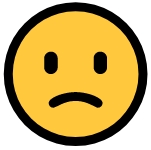}}}
\def\emojineutral{\raisebox{-0.55ex}{\includegraphics[width=1.3em]{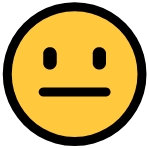}}}
\def\emojiattention{\raisebox{-0.55ex}{\includegraphics[width=1.3em]{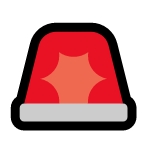}}}
\title{Can ChatGPT Understand Too? \\A Comparative  Study on ChatGPT and Fine-tuned BERT}

\author{Qihuang Zhong$^{\diamondsuit, \Re}$\thanks{~~Work was done when Qihuang was interning at JD Explore Academy.},
\ \textbf{
\ Liang Ding$^{\Re, \flat}$,
\ Juhua Liu$^{\diamondsuit}$, 
\ Bo Du$^{\diamondsuit}$,  
\ Dacheng Tao$^{\Re, \flat}$} \\
\ $^{\diamondsuit}$Wuhan University
\ $^{\Re}$JD Explore Academy, JD.com Inc.
\ $^{\flat}$ University of Sydney\\
\includegraphics[scale=0.15]{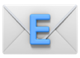} \texttt{zhongqihuang@whu.edu.cn}, \texttt{dingliang1@jd.com}\\
\includegraphics[scale=0.09]{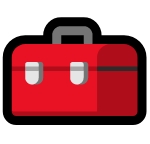} \url{https://github.com/WHU-ZQH/ChatGPT-vs.-BERT}
}

\begin{document}
\maketitle
\begin{abstract}
Recently, ChatGPT has attracted great attention, as it can generate fluent and high-quality responses to human inquiries. Several prior studies have shown that ChatGPT attains remarkable generation ability compared with existing models.
However, the quantitative analysis of ChatGPT's understanding ability has been given little attention.
In this report, we explore the understanding ability of ChatGPT by evaluating it on the most popular GLUE benchmark, and comparing it with 4 representative fine-tuned BERT-style models.
We find that: 1) ChatGPT falls short in handling paraphrase and similarity tasks; 2) ChatGPT outperforms all BERT models on inference tasks by a large margin; 3) ChatGPT achieves comparable performance compared with BERT on sentiment analysis and question-answering tasks. Additionally, by combining some advanced prompting strategies, we show that the understanding ability of ChatGPT can be further improved. 
\end{abstract}
\section{Introduction}
\label{sec:intro}
Large language models (LLMs), such as GPT-3~\cite{brown2020language} and InstructGPT~\cite{ouyangtraining}, have swept the natural language processing (NLP) community. Due to their emergent abilities~\cite{weiemergent}, these LLMs can achieve impressive few-shot and zero-shot performance in a variety of NLP tasks. More recently, ChatGPT\footnote{\url{https://chat.openai.com}}, developed by OpenAI upon InstructGPT~\cite{ouyangtraining}, has attracted great attention. Encouragingly, different from prior public chatbots, ChatGPT is able to generate fluent and comprehensive responses to various human inquiries, and even correct inappropriate human questions.

In light of the conventional wisdom that ``GPT-style models work well in generation tasks, but perform poorly for understanding tasks, even worse than the base-sized BERT~\cite{devlin2019bert}'', we wonder whether there is a similar phenomenon in the ChatGPT scenario. For the generation ability of ChatGPT, several prior studies~\cite{jiao2023chatgpt,bang2023multitask,wang23cls} have shown that ChatGPT can achieve comparable or even better performance than existing LLMs on several generation tasks. However, it is still unclear \textit{whether ChatGPT works well on natural language understanding (NLU) tasks too}.

In this report, we provide a systematic study to explore the question: ``\textit{can ChatGPT understand too}''. This question is answered by evaluating ChatGPT on the authoritative and popular GLUE~\cite{wang2018glue} benchmark, spanning 8 representative understanding tasks, i.e., sentiment analysis, linguistic acceptability, paraphrase, textual similarity, natural language inference, and question answering. For reference, we also compare it with 4 representative BERT-style models. Through a series of experiments and analyses, we find that:
\begin{itemize}
    \item[\emojisad] ChatGPT falls short in handling paraphrase and similarity tasks. Specifically, ChatGPT performs poorly in negative paraphrase and neutral similarity samples, respectively. 
    \item[\emojihappy] ChatGPT outperforms all BERT-style models on inference tasks by a large margin, indicating its impressive reasoning ability.
    \item[\emojineutral] ChatGPT achieves comparable performance compared with BERT-base on sentiment analysis and question-answering tasks.
    \item[\emojiattention] Despite its good performance on inference tasks, ChatGPT may generate some contradictory or unreasonable responses, which would be its potential limitations.
\end{itemize}

Furthermore, in addition to analyzing the ChatGPT itself, we also explore the complementarity of ChatGPT and some advanced prompting strategies, i.e., the standard few-shot prompting (also known as in-context learning)~\cite{brown2020language}, manual few-shot chain-of-thought (CoT) prompting~\cite{weichain} and zero-shot CoT prompting~\cite{kojimalarge}. Empirically, we find that \ding{182}~all these prompting strategies can consistently improve the ChatGPT, among which the manual-CoT brings the most performance benefits. Interestingly, we also observe that \ding{183}~the performance of in-context learning is relatively sensitive to the provided examples, especially in the 1-shot scenario, which is similar to the findings of~\citet{agrawal2022context}. One possible reason is that the performance of in-context learning is (highly) related to the correlation (e.g., similarity) between the provided examples and test data.

To summarize, the zero-shot performance of ChatGPT is comparable to the baseline fine-tuned BERT-base model. With the help of advanced prompting strategies, ChatGPT shows better understanding ability, and even outperforms the powerful RoBERTa-large model on some NLU tasks. However, there is still a performance gap between ChatGPT and fine-tuned RoBERTa-large in terms of average performance. That said, while ChatGPT could solve many NLP problems quite well, it still fails to beat the current SOTA models~\cite{deberta,structBERT,zhong2022superglue,turingv6,zhong2023glue}, especially on some NLU tasks. 

The remainder of this report is designed as follows. We present the evaluation settings and comparative results in Section~\ref{sec:evaluation}. In Section~\ref{sec:cot}, we explore whether ChatGPT can be improved with advanced prompting strategies. In Section~\ref{sec:related}, we briefly review the related works. Conclusions are described in Section~\ref{sec:conclusion}.
\begin{table*}[ht]
\scalebox{0.8}{
\begin{tabular}{lccclp{0.72\textwidth}}
\toprule
\textbf{Task} & \multicolumn{1}{l}{\textbf{\#Pos.}} & \multicolumn{1}{l}{\textbf{\#Neg.}} & \multicolumn{1}{l}{\textbf{\#Neu.}} & \textbf{Description} & \textbf{Template Prompt} \\ \midrule
\multicolumn{5}{c}{Single-Sentence Tasks} \\ \midrule
CoLA & 25 & 25 & - & acceptablity & \texttt{For the sentence: ``[text]", is the sentence grammarly correct?} \\ \cmidrule{6-6}
SST-2 & 25 & 25 & - & sentiment & \texttt{For the sentence: ``[text]", is the sentiment in this sentence positive or negative?} \\ \midrule
\multicolumn{5}{c}{Similarity and Paraphrase Tasks} \\ \midrule
MRPC & 25 & 25 & - & paraphrase & \texttt{For the sentence pair ``[text\_1]" and ``[text\_2]", do these two sentences have the same semantics?} \\ \cmidrule{6-6}
STS-B & \multicolumn{3}{l}{\quad total of 50} & similarity & \texttt{Determine the similarity between the following two sentences: ``[text\_1]" and ``[text\_2]". The score should be ranging from 0.0 to 5.0, and can be a decimal.} \\ \cmidrule{6-6}
QQP & 25 & 25 & - & paraphrase & \texttt{For the sentence pair ``[text\_1]" and ``[text\_2]", do these two sentences have the same semantics?} \\ \midrule
\multicolumn{5}{c}{Inference Tasks} \\ \midrule
MNLI & 25 & 25 & 25 & NLI & \texttt{Given the sentence ``[text\_1]", determine if the following statement is entailed or contradicted or neutral: ``[text\_2]"} \\ \cmidrule{6-6}
QNLI & 25 & 25 & - & QA/NLI & \texttt{Given the question ``[text\_1]", determine if the following sentence contains the corresponding answer: ``[text\_2]"} \\ \cmidrule{6-6}
RTE & 25 & 25 & - & NLI & \texttt{Given the sentence ``[text\_1]", determine if the following statement is entailed: ``[text\_2]"} \\
\bottomrule
\end{tabular}
}
\caption{Task statistics, descriptions and prompts. All tasks are single sentence or sentence pair classification, except STS-B, which is a regression task. For ease of illustration, we use ``\#Pos./\#Neg./\#Neu.'' to denote the positive, negative and neutral instances for each task.
Considering the limits of ChatGPT, we randomly sample 25 instances for each class from the dev set of each task for evaluation, except for STS-B, where we randomly sample 50 instances from a uniform distribution. In the prompts, \texttt{[text]}, \texttt{[text\_1]} and \texttt{[text\_2]} are input slots.
}
\label{tab:data}
\end{table*}

\section{ChatGPT vs. BERT}
\label{sec:evaluation}
In this section, we first introduce the evaluation setting (\S\ref{subsec:setting}), and present the major results (\S\ref{subsec:results}). Then, some analyses of why ChatGPT performs well or poorly are also provided (\S\ref{subsec:analysis}). Lastly, we show some failure examples of ChatGPT to explore its potential limitations (\S\ref{subsec:case}).

\subsection{Evaluation Setting}
\label{subsec:setting}
Here, we briefly introduce the evaluation setting, including downstream tasks and datasets, baselines, and prompts for ChatGPT.

\paragraph{Tasks and Datasets.}
Following many prior works~\cite{zhong2022e2s2,zhong2023glue}, we use the widely-used GLUE benchmark~\cite{wang2018glue} for model evaluation purposes. As one of the most popular NLU benchmarks, GLUE consists of several challenging NLU tasks, including linguistic acceptability (CoLA,~\newcite{warstadt2019neural}), sentiment analysis (SST-2,~\newcite{socher2013recursive}), paraphrase (MRPC,~\newcite{dolan2005automatically}), textual similarity (STS-B,~\newcite{cer2017semeval}), question paraphrase (QQP), textual entailment (MNLI,~\newcite{williams2018broad}, RTE,~\newcite{giampiccolo2007third}) and question-answer entailment (QNLI,~\newcite{rajpurkar2016squad}). Considering the limits of testing ChatGPT, we follow~\citet{jiao2023chatgpt} and randomly sample a subset of the dev set as the evaluation data for each task. Specifically, since most GLUE tasks are classification tasks (except STS-B which is a regression task), we randomly sample 25 instances for each class from the dev set. For STS-B, we randomly sample 50 instances from a uniform distribution. Table~\ref{tab:data} shows the task descriptions and statistics\footnote{More detailed descriptions are shown in Appendix~\ref{appendix_data}.}.

For evaluation, we report the performance with Accuracy (``\textit{Acc.}'') metric for most tasks, except the Pearson and Spearman correlation (``\textit{Pear./Spea.}'') for STS-B, the Matthew correlation (``\textit{Mcc.}'') for CoLA, the additional F1 score for MRPC and QQP.

\begin{table*}[ht]
\scalebox{0.82}{
\begin{tabular}{lccccccccccccc}
\toprule
\multicolumn{1}{c}{\multirow{2}{*}{Method}} & CoLA & SST-2 & \multicolumn{2}{c}{MRPC} & \multicolumn{2}{c}{STS-B} & \multicolumn{2}{c}{QQP} & \multicolumn{2}{c}{MNLI} & QNLI & RTE & GLUE \\ \cmidrule(lr){2-3} \cmidrule(lr){4-9} \cmidrule(lr){10-13}
\multicolumn{1}{c}{} & \textit{Mcc.} & \textit{Acc.} & \textit{Acc.} & \textit{F1} & \textit{Pear.} & \textit{Spea.} & \textit{Acc.} & \textit{F1} & \textit{m.} & \textit{mm.} & \textit{Acc.} & \textit{Acc.} & \underline{avg.} \\ \midrule
BERT-base & 56.4 & 88.0 & 90.0 & 89.8 & 83.0 & 81.9 & 80.0 & 80.0 & 82.7 & 82.7 & 84.0 & 70.0 & \underline{79.2} \\
BERT-large & 62.4 & 96.0 & 92.0 & 91.7 & 88.3 & 86.8 & 88.0 & 88.5 & 82.7 & 88.0 & 90.0 & 82.0 & \underline{85.4} \\
RoBERTa-base & 61.8 & 96.0 & 90.0 & 90.6 & 90.2 & 89.1 & 84.0 & 84.0 & 84.0 & 88.0 & 92.0 & 78.0 & \underline{84.7} \\
RoBERTa-large & 65.3 & 96.0 & 92.0 & 92.0 & 92.9 & 91.1 & 90.0 & 89.4 & 88.0 & 90.7 & 94.0 & 84.0 & \underline{87.8} \\ \midrule
ChatGPT & 56.0 & \textcolor[RGB]{0,176,80}{\textbf{92.0}} & \textcolor{red}{\textbf{66.0$^*$}} & \textcolor{red}{\textbf{72.1$^*$}} & \textcolor{red}{\textbf{80.9}} & \textcolor{red}{\textbf{72.4$^*$}} & \textcolor{red}{\textbf{78.0}} & 79.3 & \textcolor[RGB]{0,176,80}{\textbf{89.3$^*$}} & 81.3 & 84.0 & \textcolor[RGB]{0,176,80}{\textbf{88.0$^*$}} & \underline{78.7} \\
\bottomrule
\end{tabular}
}
\caption{Overall comparison between ChatGPT and fine-tuned BERT-style models on GLUE benchmark. The results in \textcolor[RGB]{0,176,80}{green} denote that ChatGPT surpasses the BERT-base model by a clear margin (> 2\% ($\uparrow$) score), while the \textcolor{red}{red} results denote ChatGPT under-performs BERT-base (> 2\% ($\downarrow$) score)). More specifically, ``*'' means that the performance difference between ChatGPT and BERT-base is larger than 10\%.}
\label{tab:main}
\end{table*}

\paragraph{Baselines.}
We compare ChatGPT (Jan 31 Version) with 4 representative BERT-style models, as the BERT models are commonly used as the baselines to evaluate the understanding ability~\cite{zhong2022panda}. Specifically, base-sized/ large-sized BERT~\cite{devlin2019bert} and RoBERTa~\cite{liu2019roberta} are used. All models are fine-tuned on the full training set for each task, where the fine-tuning hyper-parameters are the same to~\citet{zhong2022improving}. To estimate the lower bound of ChatGPT's understanding ability, we mainly focus on the comparison between ChatGPT and the basic base-sized BERT.

\begin{figure}[t]
\centering
\includegraphics[width=0.47\textwidth]{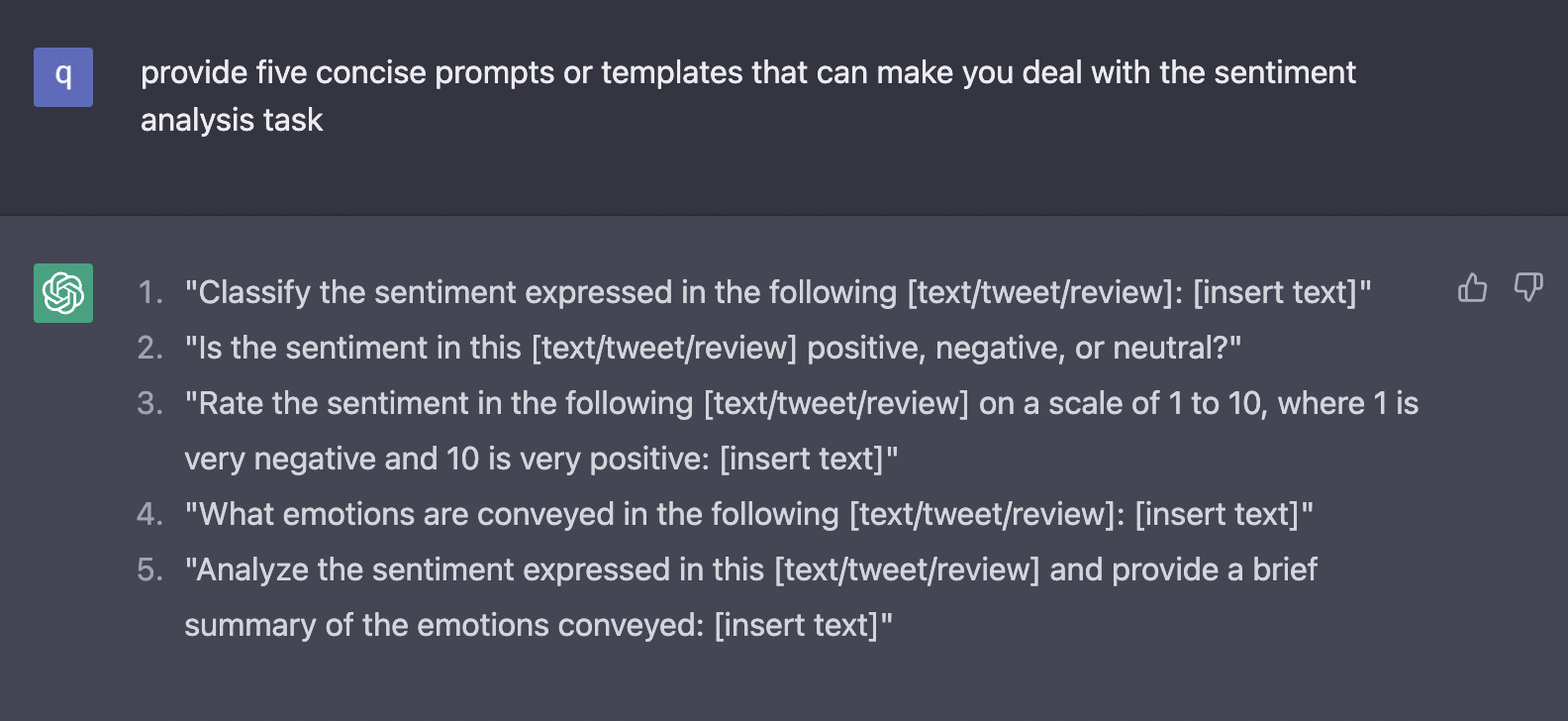}
\caption{Prompts for sentiment analysis, generated by ChatGPT.}
\label{fig:prompt}
\end{figure}

\paragraph{Prompts for ChatGPT.}
For each task, we design task-specific prompts for triggering the understanding ability of ChatGPT. Specifically, inspired by~\citet{jiao2023chatgpt}, we also ask ChatGPT to generate the prompts for each task, by inputting the following human inquiries:
\begin{itemize}
    \item[] \texttt{> provide five concise prompts or templates that can make you deal with the [x] task}
\end{itemize}
where the \texttt{[x]} is the task slot. Taking the sentiment analysis task as an example, we show this process in Figure~\ref{fig:prompt}. We evaluated ChatGPT on the sentiment analysis task with these five candidate prompts in the preliminary experiments and found a slight performance difference. Thus, for simplicity, we choose one typical prompt for each task and show them in Table~\ref{tab:data}.

\subsection{Main Results}
\label{subsec:results}
The full results on the GLUE benchmark are shown in Table~\ref{tab:main}. Overall, ChatGPT can achieve comparable average performance compared with BERT-base (78.7\% vs. 79.2\%), but still underperforms the other powerful BERT-style models (e.g., RoBERTa-large, 87.8\%) by a clear margin. These results show that ChatGPT attains the basic understanding ability, but there is still quite some room for improvement.

Specifically, comparing ChatGPT with BERT-base on specific tasks, we can find that: 1) ChatGPT performs poorly on the paraphrase and similarity tasks, i.e., MRPC and STS-B, where the performance drop is up to 24\% score. 2) ChatGPT surpasses all BERT-style models on natural language inference tasks, i.e., MNLI and RTE, indicating its superiority on inference/reasoning. 3) ChatGPT is comparable to BERT-base on the single sentence classification tasks, i.e., sentiment analysis (SST-2) and linguistic acceptability (CoLA), and QA-related tasks, i.e., QNLI.

\subsection{Analysis}
\label{subsec:analysis}
As seen in Table~\ref{tab:main}, ChatGPT works well on inference tasks, but falls short in handling paraphrase and similarity tasks. Here, we investigate how ChatGPT works on these special tasks in detail.

\begin{table*}[]
\centering
\scalebox{0.95}{
\begin{tabular}{lccccc}
\toprule
\multicolumn{1}{c}{\multirow{2}{*}{Method}} & \multicolumn{3}{c}{MNLI-m} & \multicolumn{2}{c}{RTE} \\ \cmidrule(lr){2-4} \cmidrule(lr){5-6}
\multicolumn{1}{c}{} & \multicolumn{1}{l}{Entailment} & \multicolumn{1}{l}{Contradiction} & \multicolumn{1}{l}{Neutral} & \multicolumn{1}{l}{Entailment} & \multicolumn{1}{l}{Not\_Entailment} \\ \midrule
BERT-base & 88.0 & 88.0 & 72.0 & 76.0 & 64.0 \\
BERT-large & 76.0 & 92.0 & 80.0 & 80.0 & 84.0 \\
RoBERTa-base & 84.0 & 88.0 & 80.0 & 80.0 & 76.0 \\
RoBERTa-large & 84.0 & 92.0 & 88.0 & 92.0 & 76.0 \\ \midrule
ChatGPT & 92.0$^*$ ($\uparrow$ \textcolor[RGB]{0,176,80}{\textbf{4.0}}) & 96.0$^*$ ($\uparrow$ \textcolor[RGB]{0,176,80}{\textbf{8.0}}) & 80.0 ($\uparrow$ \textcolor[RGB]{0,176,80}{\textbf{8.0}}) & 96.0$^*$ ($\uparrow$ \textcolor[RGB]{0,176,80}{\textbf{20.0}}) & 80.0 ($\uparrow$ \textcolor[RGB]{0,176,80}{\textbf{16.0}}) \\
\bottomrule
\end{tabular}
}
\caption{Per-class accuracy (\%) of ChatGPT and BERT-style models on MNLI-m and RTE. The number in parentheses indicates the
performance improvement over BERT-base. ``*'' denotes that ChatGPT outperforms all BERT-style models.}
\label{tab:nli}
\end{table*}
\begin{table}[]
\centering
\scalebox{0.95}{
\begin{tabular}{lcc}
\toprule
\multicolumn{1}{c}{\multirow{2}{*}{Method}} & \multicolumn{2}{c}{MRPC} \\ \cmidrule{2-3}
\multicolumn{1}{c}{} & \multicolumn{1}{l}{Entailment} & \multicolumn{1}{l}{Not\_Entailment} \\
\midrule
BERT-base & 88.0 & 92.0 \\
BERT-large & 88.0 & 96.0 \\
RoBERTa-base & 96.0 & 84.0 \\
RoBERTa-large & 92.0 & 92.0 \\
\midrule
ChatGPT & 88.0 ($\downarrow$ 0) & \textcolor{red}{\textbf{44.0 ($\downarrow$ 47.0)}} \\
\bottomrule
\end{tabular}
}
\caption{Per-class accuracy (\%) of ChatGPT and BERT-style models on MRPC. The number in parentheses indicates the
performance drops over BERT-base.}
\label{tab:mrpc}
\end{table}

\paragraph{Inference Tasks.} To have a closer look at why ChatGPT achieves impressive performance on inference tasks, we report the per-class accuracy of ChatGPT and compared models on MNLI and RTE tasks. The results are shown in Table~\ref{tab:nli}. It can be seen that, ChatGPT outperforms BERT-base by a large margin among all settings. Especially, in the class of ``entailment'', i.e., the premise entails the hypothesis, ChatGPT even surpasses all powerful BERT models by a clear margin. These results continue showing the effective inference ability of ChatGPT, especially reasoning factual input. 

\paragraph{Paraphrase Task.} Similar to the above analysis, we also report the per-class accuracy of ChatGPT and other models on the paraphrasing task, i.e., MRPC, in Table~\ref{tab:mrpc}. Surprisingly, ChatGPT achieves comparable performance compared with BERT-base when evaluating ``entailment'' samples, but there is a dramatic performance drop (up to 47\% score) in the class of ``not\_entailment'', where the sentences in the pair are not semantically equivalent. This indicates that ChatGPT is not sensitive to the semantic difference between a pair of sentences, which might be related to a lack of human feedback on this aspect during model training.

\begin{figure}[t]
\centering
\includegraphics[width=0.47\textwidth]{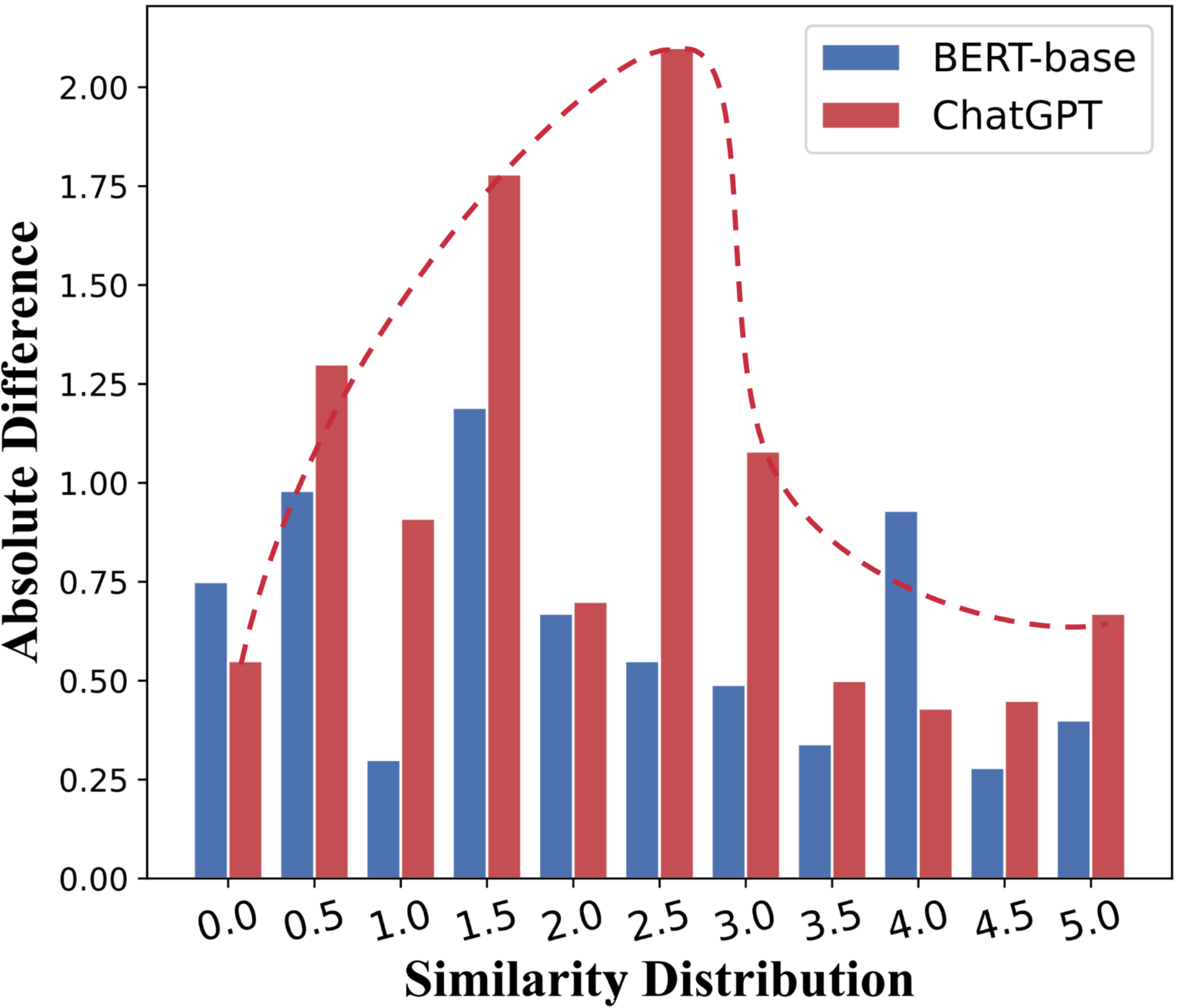}
\caption{The comparison between BERT-base and ChatGPT on STS-B. The x-axis denotes the similarity distribution of STS-B, and the y-axis denotes the absolute difference between prediction and ground truth.}
\label{fig:stsb}
\end{figure}

\paragraph{Similarity Task.} Since the STS-B is a regression task, we choose some samples from the uniform similarity distribution, ranging from 0 for no meaning overlap to 5 for meaning equivalence, and show the absolute difference between predictions and ground-truths for ChatGPT and BERT-base, respectively. As seen in Figure~\ref{fig:stsb}, ChatGPT underperforms BERT-base in most cases, as it generally predicts far from the ground-truths. To be more specific, we can observe that ChatGPT performs worse when the sentences in the pair have a lower similarity (<2.5 scores), which is similar to the observation from Table~\ref{tab:mrpc}. It can also be found that, ChatGPT is difficult to accurately predict the similarity score for a pair of sentences around the decision boundary (around the 2.5 scores). One of the reasons is ChatGPT is not fine-tuned on the STS-B task and cannot determine a correct decision boundary. And we show, in Section~\ref{sec:cot}, advanced prompting strategies upon ChatGPT could be considerably improved.

\subsection{Case Study}
\label{subsec:case}
Here, we show some bad cases of ChatGPT to explore its potential limitations, and attempt to explain why ChatGPT falls short in handling the negative samples of the paraphrasing task.

\begin{figure}[t]
\centering
\includegraphics[width=0.48\textwidth]{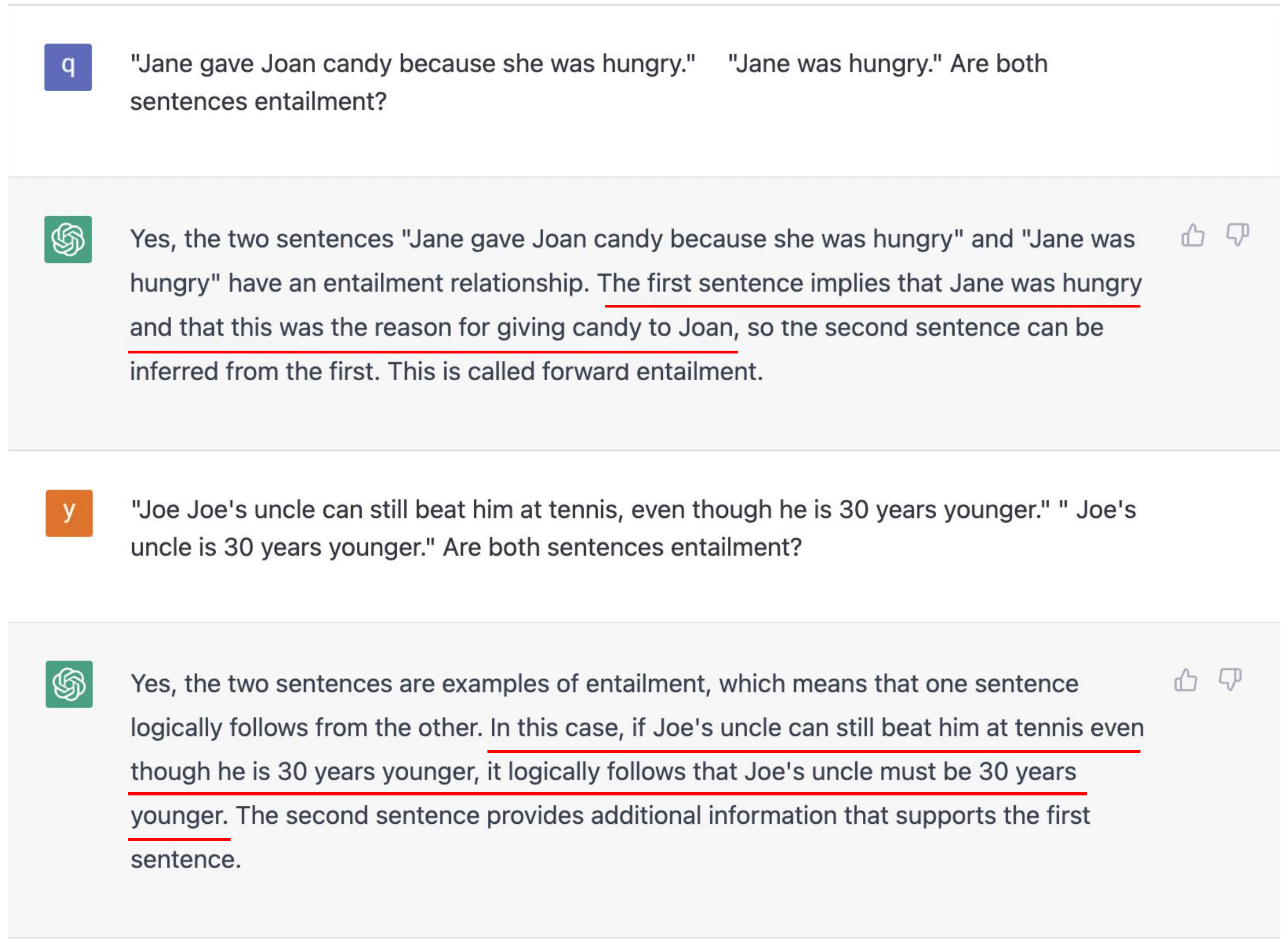}
\caption{Failures of ChatGPT in inference task. The ground truth for both cases is ``not\_entailment'', but ChatGPT makes the ``entailment'' predictions. (Data: 2022.02.09)}
\label{fig:case_nli}
\end{figure}
First, while ChatGPT works well for the inference task, it still fails to make the correct predictions in some cases. As seen in Figure~\ref{fig:case_nli}, ChatGPT can generate fluent responses to both inquiries due to its powerful generation ability. However, we observe that these responses are somewhat contradictory and even unreasonable. For example, in the upper case, ChatGPT says ``\texttt{...Jane was hungry and that this was the reason for giving candy to Joan,...}'', which is very confusing. If Jane was indeed hungry, Jane would not give candy to Joan, but eat the candy himself (herself). There is a similar phenomenon in the lower case, where ChatGPT answers with confused logic. 
In general, ChatGPT is able to generate fluent responses following a certain pattern, but appears to have limitations in really reasoning the sentences. One evidence is that ChatGPT even fails to answer the questions, such as the cases in Figure~\ref{fig:case_nli}, that are easily answered by humans.

\begin{figure}[t]
\centering
\includegraphics[width=0.48\textwidth]{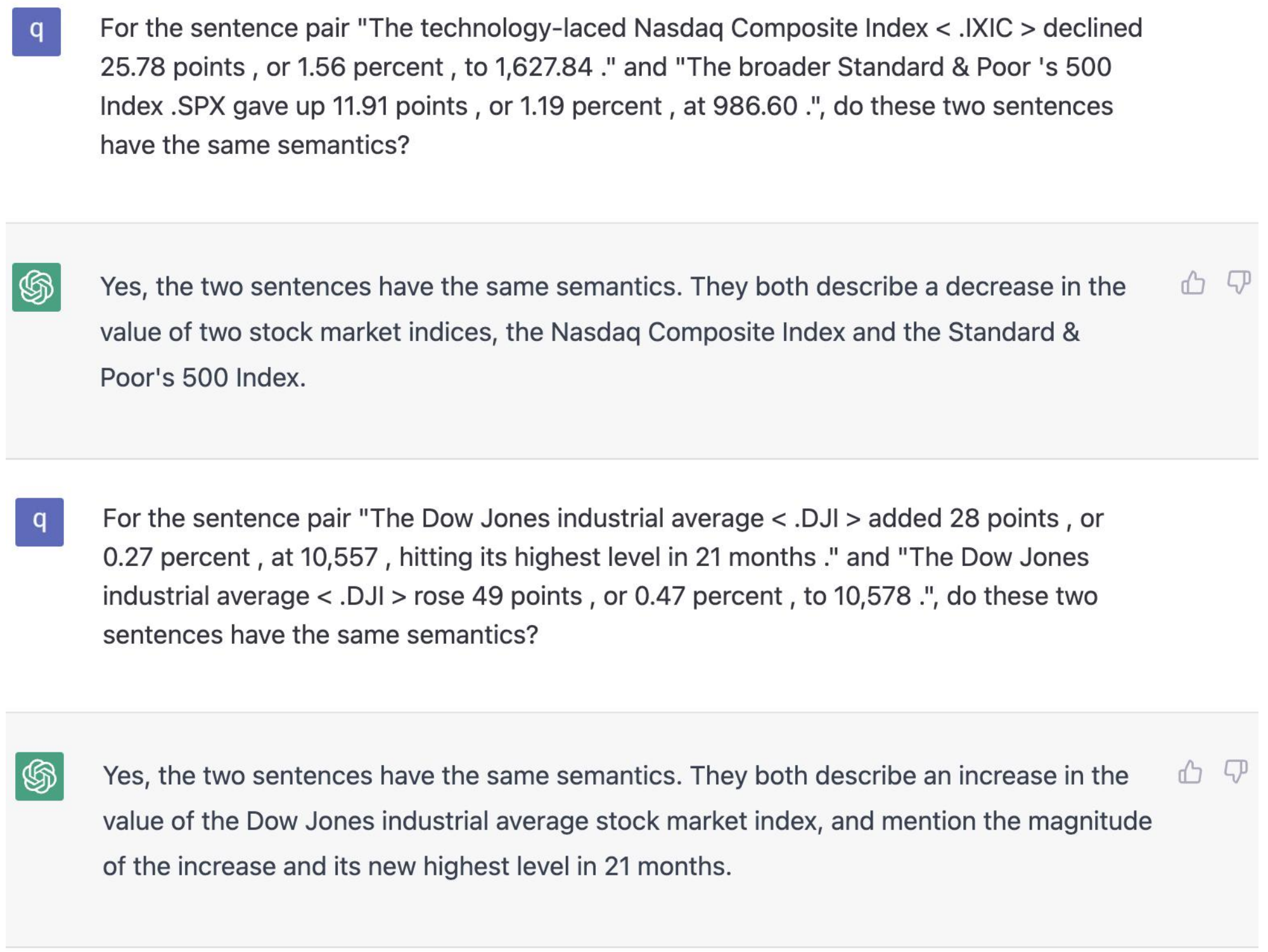}
\caption{Failures of ChatGPT in paraphrase task. The ground truth for both cases is ``not\_entailment'', but ChatGPT makes the ``entailment'' predictions. (Data: 2022.02.09)}
\label{fig:case_mrpc}
\end{figure}

\begin{figure*}[ht]
\centering
\includegraphics[width=1\textwidth]{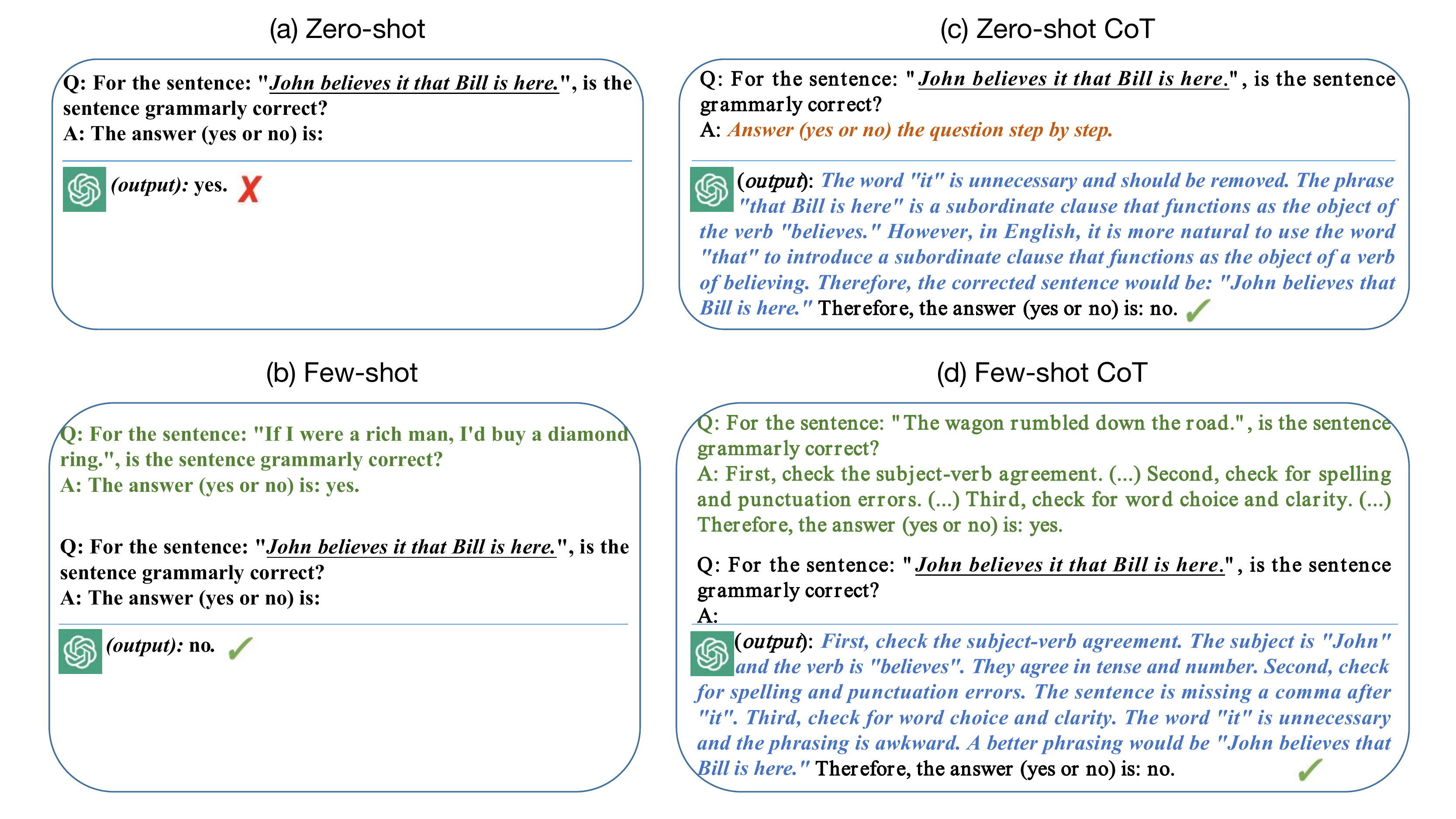}
\caption{Illustrations of ChatGPT equipped with (b) standard few-shot prompting~\cite{brown2020language}, (c) zero-shot chain-of-thought (CoT) prompting~\cite{kojimalarge} and (d) manual few-shot CoT prompting~\cite{weichain}. This test example is from the dev set of CoLA~\cite{warstadt2019neural}, while the few-shot examples (in green) are from the training set. We can find that, with the help of advanced prompting strategies, ChatGPT shows a better understanding ability.}
\label{fig:cot}
\end{figure*}

On the other hand, some example failures of ChatGPT in the paraphrase task are shown in Figure~\ref{fig:case_mrpc}. Both cases are in the class of ``not\_entailment''. ChatGPT thinks the two sentences have the same semantics, as both sentences describe a decrease (increase) in the value, which can be viewed as a coarse-grained semantic similarity. However, we can easily find that the major difference between the two sentences is the value difference, determining the ``not\_entailment'' polarity of these cases. We refer to this value difference as the fine-grained semantic difference. These cases show that such a discrepancy between coarse-grained and fine-grained semantic information might be one of the reasons why ChatGPT struggles with handling negative samples in the paraphrase task. This also indicates that strengthening the ability of ChatGPT to extract fine-grained semantic information would effectively improve its performance on the paraphrase tasks.

\section{Improving ChatGPT with Advanced Prompting Strategies}
\label{sec:cot}
As mentioned in Section~\ref{sec:evaluation}, we mainly focus on the zero-shot learning performance of ChatGPT, and the evaluation results show that there is still a clear margin between ChatGPT and fine-tuned BERT models on some NLU tasks. Inspired by some advanced prompting methods~\cite{brown2020language,weichain,kojimalarge} that can effectively exploit the capabilities of LLMs, here, we attempt to investigate whether these methods can also improve the understanding ability of ChatGPT and narrow its performance gap with powerful BERT models. 

\begin{table*}[]
\scalebox{0.82}{
\begin{tabular}{lccccccccccccc}
\toprule
\multicolumn{1}{c}{\multirow{2}{*}{Method}} & CoLA & SST-2 & \multicolumn{2}{c}{MRPC} & \multicolumn{2}{c}{STS-B} & \multicolumn{2}{c}{QQP} & \multicolumn{2}{c}{MNLI} & QNLI & RTE & GLUE \\ \cmidrule(lr){2-3} \cmidrule(lr){4-9} \cmidrule(lr){10-13}
\multicolumn{1}{c}{} & \textit{Mcc.} & \textit{Acc.} & \textit{Acc.} & \textit{F1} & \textit{Pear.} & \textit{Spea.} & \textit{Acc.} & \textit{F1} & \textit{m.} & \textit{mm.} & \textit{Acc.} & \textit{Acc.} & \underline{avg.} \\ \midrule
BERT-base & 56.4 & 88.0 & 90.0 & 89.8 & 83.0 & 81.9 & 80.0 & 80.0 & 82.7 & 82.7 & 84.0 & 70.0 & \underline{79.2} \\
RoBERTa-large & 65.3 & 96.0 & \textbf{92.0} & \textbf{92.0} & \textbf{92.9} & \textbf{91.1} & \textbf{90.0} & 89.4 & 88.0 & \textbf{90.7} & \textbf{94.0} & 84.0 & \underline{87.8} \\ \midrule
ChatGPT & 56.0 & 92.0 & 66.0 & 72.1 & 80.9 & 72.4 & 78.0 & 79.3 & \textbf{89.3} & 81.3 & 84.0 & 88.0 & \underline{78.7} \\ \hdashline
\multicolumn{14}{l}{\textit{Standard few-shot prompting}~\cite{brown2020language}} \\
\quad -w/ 1-shot & 52.0 & 96.0 & 66.0 & 65.3 & 87.4 & 87.0 & 84.0 & 83.3 & 80.0 & 78.7 & 84.0 & 80.0 & \underline{78.5} \\
\quad -w/ 5-shot & 60.2 & \textbf{98.0} & 76.0 & 77.8 & 89.0 & 86.9 & 90.0 & 89.8 & 82.7 & 84.0 & 88.0 & 86.0 & \underline{83.8} \\
\hdashline
\multicolumn{14}{l}{\textit{Zero-shot CoT}~\cite{kojimalarge}} \\
\quad -w/ zero-shot CoT & 64.5 & 96.0 & 78.0 & 76.6 & 87.1 & 87.8 & 80.0 & 80.8 & 86.7 & 89.3 & 86.0 & 90.0 & \underline{83.7} \\ \hdashline
\multicolumn{14}{l}{\textit{Manual few-shot CoT}~\cite{weichain}} \\
\quad -w/ 1-shot CoT & 60.8 & 94.0 & 82.0 & 83.2 & 89.1 & 88.7 & 84.0 & 82.6 & 85.3 & 84.0 & 88.0 & \textbf{92.0} & \underline{84.3} \\
\quad -w/ 5-shot CoT & \textbf{68.2} & 96.0 & 82.0 & 81.6 & 90.0 & 90.2 & 86.0 & 85.1 & 85.3 & 86.7 & 90.0 & \textbf{92.0} & \underline{86.2} \\
\bottomrule
\end{tabular}
}
\caption{Results of ChatGPT equipped with advanced prompting strategies. For reference, we also report the results of baseline BERT-base and powerful RoBERTa-large. The best results are in \textbf{bold}. We can find that all advanced prompting strategies bring some performance improvements to ChatGPT, among which the manual few-shot CoT is empirically optimal.}
\label{tab:cot_result}
\end{table*}
\subsection{Advanced Prompting Strategies}
In this study, we use three popular prompting strategies as follows: 
\begin{itemize}
    \item \textbf{Standard few-shot prompting}: also known as in-context learning~\cite{brown2020language}, it can simply ``prompt'' the model with a few input-output exemplars demonstrating the task. Specifically, as shown in Figure~\ref{fig:cot} (b), it enables the ChatGPT to perform a target task by feeding a few prompted examples as part of the input.
    \item \textbf{Manual few-shot CoT prompting}: chain-of-thought (CoT) prompting is proposed by~\citet{weichain}, which provides manual intermediate reasoning steps (demonstrations)\footnote{The human efforts in the design of these demonstrations for different tasks are nontrivial. In our experience, we can first ask ChatGPT to generate the steps to perform the target task, and manually modify the generated reasoning steps. After obtaining one demonstration, we can encourage the ChatGPT to generate similar demonstrations for other input examples.} to lead the model to output the final answer step by step. 
    \item \textbf{Zero-shot CoT}: instead of manually designing the demonstrations, \citet{kojimalarge} propose a zero-shot CoT method, which employs a simple and straightforward template-based prompting for CoT reasoning. Specifically, as shown in Figure~\ref{fig:cot} (c), we use ``\texttt{Answer (yes or no) the question step by step.}'' to extract step-by-step reasoning.
\end{itemize}

To have a close look, taking the CoLA task as an example, we show the illustrations of ChatGPT equipped with these prompting strategies in Figure~\ref{fig:cot}. More input examples for each task can be found in Appendix~\ref{appendix_cot}.

\subsection{More Results and Analyses}
The overall results of ChatGPT equipped with advanced prompting strategies on GLUE benchmark are shown in Table~\ref{tab:cot_result}. For reference, we also compare the improved ChatGPT with the baseline BERT-base and powerful RoBERTa-large models. Based on these empirical results, we can further find that:

\paragraph{\ding{182} ChatGPT benefits from all these prompting strategies.}
Compared to the baseline ChatGPT (78.7\%), i.e., zero-shot ChatGPT, all these prompting strategies bring some performance improvements. Specifically, the standard few-shot prompting and zero-shot CoT improves the overall performance of ChatGPT by +5.1\% and +5.0\% average scores, respectively. More encouragingly, with the help of manual few-shot CoT, ChatGPT achieves up to +7.5\% average gains and even outperforms most BERT-style models (except for RoBERTa-large). These results indicate that prompting the ChatGPT with manual-CoT could be the Pareto frontier for leveraging its capabilities. 

\begin{figure}[t]
\centering
\includegraphics[width=0.47\textwidth]{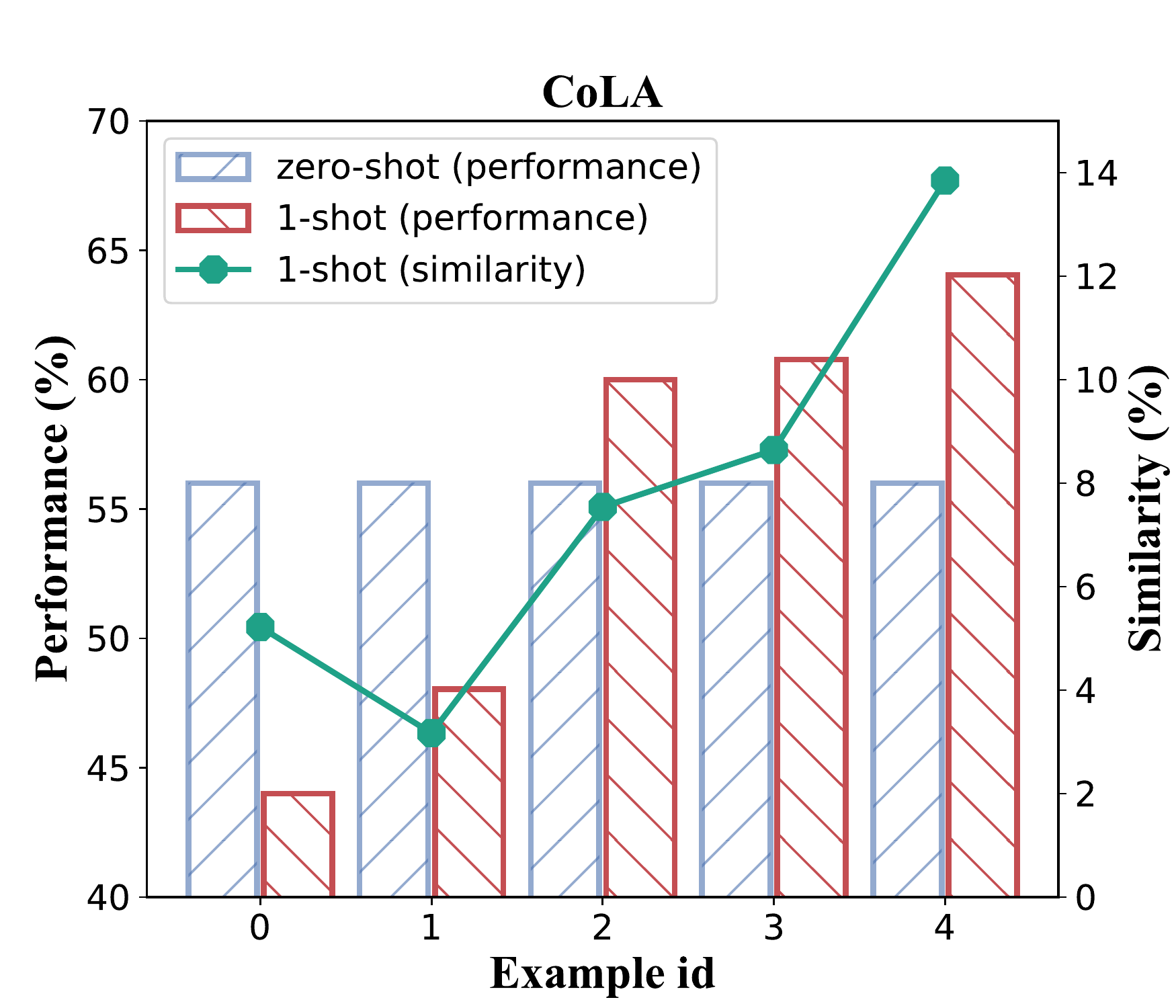}
\caption{Analysis of the unstable 1-shot prompting performance on the CoLA task. The x-axis denotes 5 randomly sampled examples. The left y-axis is the performance of ChatGPT, while the right y-axis is the average textual similarity, measured by Sentence-BERT~\cite{reimers-2019-sentence-bert}, between the given example and test data. 
}
\label{fig:1-shot}
\end{figure}

\paragraph{\ding{183} In the 1-shot scenario, the performance of ChatGPT is relatively sensitive to the given in-context example.}
Despite the overall performance gains in few-shot settings, we can find that ChatGPT does not consistently perform better on these NLU tasks, especially in the 1-shot scenario. More specifically, when equipped with the standard 1-shot prompting, ChatGPT even performs worse on some tasks, e.g., CoLA, MRPC, MNLI and RTE. We attribute it to the lower correlation between the randomly sampled in-context example and test data, as the prior work~\cite{agrawal2022context} shows that the 1-shot noisy unrelated example could have a catastrophic impact on output quality\footnote{This might be also the reason why 5-shot prompting generally works better, as concatenating multiple random examples could reduce the effect of noise.}. To further verify this conjecture, we use the different 1-shot example to perform the standard 1-shot prompting. Taking the CoLA task as an example, the comparative results are shown in Figure~\ref{fig:1-shot}. As seen, the 1-shot performance is unstable, and when given a more related 1-shot example, ChatGPT can achieve more performance gains, confirming our statement.

\paragraph{\ding{184} There is still a performance gap between ChatGPT and fine-tuned RoBERTa-large.} 
With the help of manual-CoT, ChatGPT achieves impressive performance improvements and shows state-of-the-art (SOTA) performance among all comparison models on some tasks, e.g., CoLA, SST-2 and RTE. However, as seen, compared with the fine-tuned RoBERTa-large, ChatGPT still underperforms on some tasks, especially for the paraphrase task (MRPC), by a clear margin. These results continue indicating that, although ChatGPT could solve many NLP problems quite well, it still fails to beat the current SOTA models, especially on some NLU tasks.

\paragraph{\ding{43} Note} Some readers may concern that our work could be a kind of ``lottery ticket'', as we only evaluate ChatGPT on a part of the validation set for each task. To dispel such doubt, we investigate whether there are similar findings in the full-data setting. Specifically, taking the RTE task as an example, we report the corresponding results of ChatGPT under the few-data and full-data settings, respectively, as shown in Table~\ref{tab:full_data_result}. It can be found that ChatGPT shows similar characteristics (e.g., significantly benefiting from manual-CoT) in both scenarios, indicating the credibility of our work. 

\begin{table}[]
\centering
\scalebox{0.9}{
\begin{tabular}{lcc}
\toprule
\multicolumn{1}{c}{\multirow{1}{*}{Method}} & Few-data & Full-data \\ \midrule
ChatGPT & 88.0 & 83.8 \\ \hdashline
\multicolumn{3}{l}{\textit{Standard few-shot prompting}} \\
\quad -w/ 1-shot & 80.0 & 83.4  \\
\quad -w/ 5-shot & 86.0 & 84.4  \\
\hdashline
\multicolumn{3}{l}{\textit{Zero-shot CoT}} \\
\quad -w/ zero-shot CoT & 90.0 & 85.9  \\ \hdashline
\multicolumn{3}{l}{\textit{Manual few-shot CoT}} \\
\quad -w/ 1-shot CoT & 92.0 & 87.0  \\
\quad -w/ 5-shot CoT & 92.0 & 89.9 \\
\bottomrule
\end{tabular}
}
\caption{Results of ChatGPT evaluated on the few-data(the setting used in our main experiment)/ full data of RTE task. We can find that there are similar findings in both scenarios.}
\label{tab:full_data_result}
\end{table}
\section{Related Works}
\label{sec:related}
In recent years, we have witnessed numerous Transformer-based pretrained language models (PLMs)~\cite{devlin2019bert,liu2019roberta,brown2020language,raffel2020exploring,lewis2020bart,zhong2022e2s2,zhong2023glue} that achieved tremendous success in various natural language processing (NLP) tasks. Based on the model architectures, these PLMs can be classified into three groups: 1) encoder-only PLMs (e.g., BERT~\cite{devlin2019bert})\footnote{We refer to these encoder-only models as BERT-style models, and the decoder-only models as GPT-style models.}, 2) decoder-only PLMs (e.g., GPT-3~\cite{brown2020language}) and 3) encoder-decoder PLMs (e.g., T5~\cite{raffel2020exploring}). Due to different pretraining functions, these PLMs exhibit different abilities when performing NLP tasks. Specifically, the BERT-style models are based on a bidirectional masked language modeling (MLM) objective, which enforces the models to encode the context information. Through fine-tuning on the specific task, these BERT-style models can work well on a variety of natural language understanding (NLU) tasks. On the contrary, the GPT-style models aim to predict future words towards a sequence of words. Such auto-regressive models are well-suitable for language generation, but they are unidirectional and usually fail short in the representation learning for understanding the sentence~\cite{liu2021pre, zhong2022e2s2}. 

More recently, a lot of work focus on scaling up the PLMs and developing the large language models (LLMs)~\cite{ouyangtraining,chowdhery2022palm,smith2022using,zhang2022opt}. \citet{weiemergent} show that LLMs exhibit emergent abilities, e.g., few-shot and zero-shot learning, when the model sizes are large enough. As a typical LLM, the recently-released ChatGPT has attracted great attention, due to its impressive ability to generate fluent and high-quality responses. There is growing interest in exploring the capabilities, applications, ethics, and failures of ChatGPT~\cite{jiao2023chatgpt,bang2023multitask,qin2023chatgpt,zhuo2023exploring,wang23cls}. Along with the research line, we mainly focus on analyzing the understanding ability of ChatGPT in this report, which is important but has been given little attention. 
\section{Conclusion}
\label{sec:conclusion}
In this study, we empirically investigate the language understanding ability of ChatGPT on a diversity of natural language understanding tasks. Through a series of quantitative studies, we find that ChatGPT works well on inference tasks, but falls short in handling paraphrase and similarity tasks, especially for the negative instances. Furthermore, we attempt to improve the understanding ability of ChatGPT with some advanced prompting strategies. The results show that with the help of these prompting strategies, ChatGPT can achieve significant performance improvements, and even outperforms the powerful RoBERTa-large on some tasks.
Overall, ChatGPT attains a comparable understanding ability compared with some fine-tuned BERT-style models, but still fails to beat the currently best models on some NLU tasks. We hope our study could facilitate more research on how to address the limitations and improve the understanding performance of ChatGPT.

\section*{Limitations}
Our work has several potential limitations.
First, due to the limits of testing ChatGPT, we mainly evaluate ChatGPT on a part of the validation set for each task. It would be more convincing if we can test on more samples. On the other hand, this report only uses the GLUE benchmark for experiments, in which the task types are somewhat limited. In future work, we would like to evaluate ChatGPT on more NLU tasks and conduct more in-depth analyses and discussions. 

\bibliography{acl2023}
\bibliographystyle{acl_natbib}

\appendix
\section{Appendix}
\label{sec:appendix}

\subsection{Details of Tasks}
\label{appendix_data}
In this work, we conduct extensive experiments on the GLUE~\cite{wang2018glue} benchmark. Here, we introduce the detailed descriptions of all downstream tasks and datasets as follows:

\textbf{CoLA} Corpus of Linguistic Acceptability~\cite{warstadt2019neural} is a binary single-sentence classification task to determine whether a given sentence is linguistically ``acceptable''.

\textbf{SST-2} The Stanford Sentiment Treebank~\cite{socher2013recursive} is a binary classification task to predict the sentiment of a given
sentence.

\textbf{MRPC} Microsoft Research Paraphrase Corpus~\cite{dolan2005automatically} is a task to predict whether two sentences are semantically equivalent.

\textbf{STS-B} Semantic Textual Similarity~\cite{cer2017semeval} is a task to predict how similar two sentences are on a 1-5 scale in terms of semantic meaning.

\textbf{QQP} The Quora Question Pairs dataset is a collection of question pairs from the community question-answering website Quora. The task is to determine whether a pair of questions are semantically equivalent. 

\textbf{MNLI} The Multi-Genre Natural Language Inference Corpus~\cite{williams2018broad} is a task to predict whether the premise entails the hypothesis, contradicts the hypothesis, or neither, given a premise sentence and a hypothesis sentence.

\textbf{QNLI} Question Natural Language Inference is a binary classification task constructed from SQuAD~\cite{rajpurkar2016squad}, which aims to predict whether a context sentence contains the answer to a question sentence. 

\textbf{RTE} Recognizing Textual Entailment~\cite{giampiccolo2007third}, given a premise and a hypothesis, is a task to predict whether the premise entails the hypothesis. 

\subsection{Input Examples}
\label{appendix_cot}
Here, we present input examples of standard few-shot prompting, zero-shot CoT prompting and manual few-shot CoT prompting used in ChatGPT. Table~\ref{tab_cola} to~\ref{tab_rte} show the detailed examples for each task of the GLUE benchmark.

\begingroup
\begin{table*}[htb]
    \centering
    \caption{
    Examples of standard few-shot prompting, zero-shot CoT prompting and manual few-shot CoT prompting produced by ChatGPT on the CoLA task. 
    }
    \vspace{2.8mm}
    \begin{tabular}{p{0.96\linewidth}}
    \toprule
\textit{\ding{182}~Standard few-shot prompting} \\
\vspace{-1mm}
\textbf{Q:} For the sentence: "The wagon rumbled down the road.", is the sentence grammarly correct? \\
\textbf{A:} The answer (yes or no) is: yes. \\
\vspace{-1mm}
\textbf{Q:} For the sentence: "Surprised me that you came early.", is the sentence grammarly correct? \\
\textbf{A:} The answer (yes or no) is: no. \\ \hdashline
\vspace{-1mm}
\textbf{Q:} For the sentence: "\textit{Out of the box jumped a little white rabbit.}", is the sentence grammarly correct? \\
\textbf{A:} The answer (yes or no) is: \_\_\_\_ \\ \midrule

\textit{\ding{183}~Zero-shot CoT prompting} \\
\vspace{-1mm}
\textbf{Q:} For the sentence: "\textit{Out of the box jumped a little white rabbit.}", is the sentence grammarly correct? \\
\textbf{A:} \textbf{\textit{Answer (yes or no) the question step by step.}} The answer is: \_\_\_\_ \\ \midrule
\textit{\ding{184}~Manual few-shot CoT prompting} \\
\vspace{-1mm}
\textbf{Q:} For the sentence: "The wagon rumbled down the road.", is the sentence grammarly correct?  \\
\textbf{A:} First, check the subject-verb agreement. The subject is “wagon”, and the verb is “rumbled”. They agree in tense and number. Second, check for spelling and punctuation errors. There are no spelling errors and the sentence is punctuated correctly. Third, check for word choice and clarity. The word choice is appropriate and clear. Therefore, we can conclude that the sentence is grammatically correct. The answer (yes or no) is: yes. \\
\vspace{-1mm}
\textbf{Q:} For the sentence: "Surprised me that you came early.", is the sentence grammarly correct? \\
\textbf{A:} First, check the subject-verb agreement. The subject is “you”, and there is no verb in the sentence. This makes the sentence incomplete and ungrammatical. Second, check for spelling and punctuation errors. There are no spelling errors, but the sentence is missing a comma after “Surprised me”. Third, check for word choice and clarity. The word choice is awkward and unclear. A better phrasing would be “I was surprised that you came early.” Therefore, we can conclude that the sentence is not grammatically correct. The answer (yes or no) is: no. \\ \hdashline
\vspace{-1mm}
\textbf{Q:} For the sentence: "\textit{Out of the box jumped a little white rabbit.}", is the sentence grammarly correct? \\
\textbf{A:} The answer (yes or no) is: \_\_\_\_ \\ 
\bottomrule
\end{tabular}
\label{tab_cola}
\end{table*}
\endgroup

\begingroup
\begin{table*}[htb]
    \centering
    \caption{
    Examples of standard few-shot prompting, zero-shot CoT prompting and manual few-shot CoT prompting produced by ChatGPT on the SST-2 task. 
    }
    \vspace{2.8mm}
    \begin{tabular}{p{0.96\linewidth}}
    \toprule
\textit{\ding{182}~Standard few-shot prompting} \\
\vspace{-1mm}
\textbf{Q:} For the sentence: "'ll keep you wide awake and ... very tense ", is the sentiment in this sentence positive or negative? \\
\textbf{A:} The answer (positive or negative) is: positive. \\
\vspace{-1mm}
\textbf{Q:} For the sentence: "is to see two academy award winning actresses ( and one academy award winning actor ) succumb to appearing in this junk that 's tv sitcom material at best . ", is the sentiment in this sentence positive or negative? \\
\textbf{A:} The answer (positive or negative) is: negative. \\ \hdashline
\vspace{-1mm}
\textbf{Q:} For the sentence: "\textit{entertains by providing good , lively company . }", is the sentiment in this sentence positive or negative? \\
\textbf{A:} The answer (positive or negative) is: \_\_\_\_ \\ \midrule

\textit{\ding{183}~Zero-shot CoT prompting} \\
\vspace{-1mm}
\textbf{Q:} For the sentence: "\textit{entertains by providing good , lively company . }", is the sentiment in this sentence positive or negative? \\
\textbf{A:} \textbf{\textit{Answer (positive or negative) the question step by step.}} The answer is: \_\_\_\_ \\ \midrule
\textit{\ding{184}~Manual few-shot CoT prompting} \\
\vspace{-1mm}
\textbf{Q:} For the sentence: ", while it may not rival the filmmaker 's period pieces , is still very much worth seeing ", is the sentiment in this sentence positive or negative?  \\
\textbf{A:} First, identify the main clause and its subject and verb. Main clause: "is still very much worth seeing". Subject: "it". Verb: "is". Second, determine the tone of the main clause. The tone of the main clause is positive because it states that "it is still very much worth seeing.". Third, determine the tone of the entire sentence. The sentence acknowledges that the film may not be as good as the filmmaker's other works, but still recommends it as being worth seeing. The tone of the entire sentence is somewhat positive. Therefore, we can conclude that the sentiment of this sentence is positive. The answer (positive or negative) is: positive. \\
\vspace{-1mm}
\textbf{Q:} For the sentence: "anyone could desire to the cheap , graceless , hackneyed sci-fi serials of the '30s and '40s ", is the sentiment in this sentence positive or negative?  \\
\textbf{A:} First, identify the main clause and its subject and verb. Main clause: "anyone could desire". Subject: "anyone". Verb: "could desire". Second, determine the tone of the main clause. The tone of the main clause is negative because it states that "anyone could desire." Third, determine the tone of the entire sentence. The sentence is negative overall because it implies that it would be undesirable for anyone to desire the described sci-fi serials. Therefore, we can conclude that the sentiment of this sentence is negative. The answer (positive or negative) is: negative. \\ \hdashline
\vspace{-1mm}
\textbf{Q:} For the sentence: "\textit{entertains by providing good , lively company . }", is the sentiment in this sentence positive or negative? \\
\textbf{A:} The answer (positive or negative) is: \_\_\_\_ \\ 

\bottomrule
    \end{tabular}
    \label{tab_sst2}
\end{table*}
\endgroup

\begingroup
\begin{table*}[htb]
    \centering
    \caption{
    Examples of standard few-shot prompting, zero-shot CoT prompting and manual few-shot CoT prompting produced by ChatGPT on the MRPC task. 
    }
    \vspace{2.8mm}
    \begin{tabular}{p{0.96\linewidth}}
    \toprule
\textit{\ding{182}~Standard few-shot prompting} \\
\vspace{-1mm}
\textbf{Q:} For the sentence pair "He found that men who had ejaculated more than five times a week in their 20s were a third less likely to develop aggressive prostate cancer later in life ." and "Those who ejaculated more than five times a week were a third less likely to develop serious prostate cancer in later life .", do these two sentences have the same semantics? \\
\textbf{A:} The answer (yes or no) is: yes. \\
\vspace{-1mm}
\textbf{Q:} For the sentence pair "Analysts say Davis , who faces a historic recall election in October , could get a boost in the polls with a budget plan in place ." and "Analysts say Davis , a Democrat , could get a boost in the polls if the 29-day-old budget crisis is resolved without further delay .", do these two sentences have the same semantics? \\
\textbf{A:} The answer (yes or no) is: no. \\ \hdashline
\vspace{-1mm}
\textbf{Q:} For the sentence pair "\textit{Terri Schiavo , 39 , underwent the procedure at the Tampa Bay area hospice where she has been living for several years , said her father , Bob Schindler .}" and "\textit{The tube was removed Wednesday from Terri Schiavo , 39 , at the Tampa Bay-area hospice where she has lived for several years .}", do these two sentences have the same semantics? \\
\textbf{A:} The answer (yes or no) is: \_\_\_\_ \\ \midrule

\textit{\ding{183}~Zero-shot CoT prompting} \\
\vspace{-1mm}
\textbf{Q:} For the sentence pair "\textit{Terri Schiavo , 39 , underwent the procedure at the Tampa Bay area hospice where she has been living for several years , said her father , Bob Schindler .}" and "\textit{The tube was removed Wednesday from Terri Schiavo , 39 , at the Tampa Bay-area hospice where she has lived for several years .}", do these two sentences have the same semantics? \\
\textbf{A:} \textbf{\textit{Answer (yes or no) the question step by step.}} The answer is: \_\_\_\_ \\ \midrule

\textit{\ding{184}~Manual few-shot CoT prompting} \\
\vspace{-1mm}
\textbf{Q:} For the sentence pair "He found that men who had ejaculated more than five times a week in their 20s were a third less likely to develop aggressive prostate cancer later in life ." and "Those who ejaculated more than five times a week were a third less likely to develop serious prostate cancer in later life .", do these two sentences have the same semantics?  \\
\textbf{A:} First, identify the key differences between the two sentences. Second, consider the impact of the difference in wording. Third, consider the overall meaning of the two sentences. Therefore, given that the two sentences convey the same general idea, despite the difference in wording, we can conclude that they have the same semantics. The answer (yes or no) is: yes. \\
\vspace{-1mm}
\textbf{Q:} For the sentence pair "Sen. Bob Graham , Florida Democrat , raised \$ 2 million after getting a late start ." and "Further back , Sen. Bob Graham of Florida reported about \$ 1.7 million on hand .", do these two sentences have the same semantics?  \\
\textbf{A:} First, identify the key differences between the two sentences. Second, consider the impact of the difference in wording. Third, consider the overall meaning of the two sentences. While both sentences are about Sen. Graham's financial situation, they are focused on different aspects of it and do not convey the same information. Therefore, we can conclude that the two sentences do not have the same semantics. The answer (yes or no) is: no. \\ \hdashline
\vspace{-1mm}
\textbf{Q:} For the sentence pair "\textit{Terri Schiavo , 39 , underwent the procedure at the Tampa Bay area hospice where she has been living for several years , said her father , Bob Schindler .}" and "\textit{The tube was removed Wednesday from Terri Schiavo , 39 , at the Tampa Bay-area hospice where she has lived for several years .}", do these two sentences have the same semantics? \\
\textbf{A:} The answer (yes or no) is: \_\_\_\_ \\ 

\bottomrule
    \end{tabular}
    \label{tab_mrpc}
\end{table*}
\endgroup

\begingroup
\begin{table*}[htb]
    \centering
    \caption{
    Examples of standard few-shot prompting, zero-shot CoT prompting and manual few-shot CoT prompting produced by ChatGPT on the STS-B task. 
    }
    \vspace{2.8mm}
    \begin{tabular}{p{0.96\linewidth}}
    \toprule
\textit{\ding{182}~Standard few-shot prompting} \\
\vspace{-1mm}
\textbf{Q:} Determine the similarity between the following two sentences: "'Scores of bodies' found in Syria" and "Eight more bodies found on ship". The score should be ranging from 0.0 to 5.0, and can be a decimal. \\
\textbf{A:} The answer (decimals ranging from 0 to 5) is: 1.4 \\
\vspace{-1mm}
\textbf{Q:} Determine the similarity between the following two sentences: "The man cut some carpet with a knife." and "A man is cutting carpet with a knife.". The score should be ranging from 0.0 to 5.0, and can be a decimal. \\
\textbf{A:} The answer (decimals ranging from 0 to 5) is: 4.750 \\ \hdashline
\vspace{-1mm}
\textbf{Q:} Determine the similarity between the following two sentences: "\textit{At least 38 Morsi supporters die in clashes}" and "\textit{Dozens of Morsi supporters killed in Egypt clashes}". The score should be ranging from 0.0 to 5.0, and can be a decimal. \\
\textbf{A:} The answer (decimals ranging from 0 to 5) is: \_\_\_\_ \\ \midrule

\textit{\ding{183}~Zero-shot CoT prompting} \\
\vspace{-1mm}
\textbf{Q:} Determine the similarity between the following two sentences: "\textit{At least 38 Morsi supporters die in clashes}" and "\textit{Dozens of Morsi supporters killed in Egypt clashes}". The score should be ranging from 0.0 to 5.0, and can be a decimal. \\
\textbf{A:} \textbf{\textit{Answer (ranging from 0 to 5) the question step by step.}} The answer is: \_\_\_\_ \\
\midrule

\textit{\ding{184}~Manual few-shot CoT prompting} \\
\vspace{-1mm}
\textbf{Q:} Determine the similarity between the following two sentences: "Ernst \& Young has denied any wrongdoing and plans to fight the allegations." and "Ernst \& Young has denied the SEC's claims, and called its recommendations "irresponsible".". The score should be ranging from 0.0 to 5.0, and can be a decimal.  \\
\textbf{A:} First, preprocessing the two sentences to make them more suitable for comparison. We can do this using the NLTK toolkit. Second, choosing a similarity metric. Here, we use the popular cosine similarity metric. Third, computing the similarity score between two preprocessed sentence. After obtaining the cosine similarity score, we simply multiply it by 5 to convert the similarity score to a 0.0 to 5.0 scale. Based on these, the answer (ranging from 0 to 5) is: 3.0 \\
\vspace{-1mm}
\textbf{Q:} Determine the similarity between the following two sentences: "I'm not outraged, I'm disappointed." and "God bless I'm not outraged, I'm disappointed.". The score should be ranging from 0.0 to 5.0, and can be a decimal.  \\
\textbf{A:} First, preprocessing the two sentences to make them more suitable for comparison. We can do this using the NLTK toolkit. Second, choosing a similarity metric. Here, we use the popular cosine similarity metric. Third, computing the similarity score between two preprocessed sentences. After obtaining the cosine similarity score, we simply multiply it by 5 to convert the similarity score to a 0.0 to 5.0 scale. When we preprocess the two sentences and compute their cosine similarity score, we get a score of 1.0. Multiplying this score by 5 gives us a similarity score of 5. Therefore, the similarity score between the two sentences is 5 out of 5.0. \\ \hdashline
\vspace{-1mm}
\textbf{Q:} Determine the similarity between the following two sentences: "\textit{At least 38 Morsi supporters die in clashes}" and "\textit{Dozens of Morsi supporters killed in Egypt clashes}". The score should be ranging from 0.0 to 5.0, and can be a decimal. \\
\textbf{A:} The answer (decimals ranging from 0 to 5) is: \_\_\_\_ \\
\bottomrule
    \end{tabular}
    \label{tab_stsb}
\end{table*}
\endgroup

\begingroup
\begin{table*}[htb]
    \centering
    \caption{
    Examples of standard few-shot prompting, zero-shot CoT prompting and manual few-shot CoT prompting produced by ChatGPT on the QQP task. 
    }
    \vspace{2.8mm}
    \begin{tabular}{p{0.96\linewidth}}
    \toprule
\textit{\ding{182}~Standard few-shot prompting} \\
\vspace{-1mm}
\textbf{Q:} For the sentence pair "Who is winning the presidential election, Trump or Clinton?" and "Who's winning the election, Trump or Clinton?", do these two sentences have the same semantics? \\
\textbf{A:} The answer (yes or no) is: yes. \\
\vspace{-1mm}
\textbf{Q:} For the sentence pair "How should ask my crush for her number?" and "What should I ask my crush?", do these two sentences have the same semantics? \\
\textbf{A:} The answer (yes or no) is: no. \\ \hdashline
\vspace{-1mm}
\textbf{Q:} For the sentence pair "\textit{What are your favorite photos of yourself?}" and "\textit{Which is your favourite picture of yourself?}", do these two sentences have the same semantics? \\
\textbf{A:} The answer (yes or no) is: \_\_\_\_ \\ \midrule

\textit{\ding{183}~Zero-shot CoT prompting} \\
\vspace{-1mm}
\textbf{Q:} For the sentence pair "\textit{What are your favorite photos of yourself?}" and "\textit{Which is your favourite picture of yourself?}", do these two sentences have the same semantics? \\
\textbf{A:} \textbf{\textit{Answer (yes or no) the question step by step.}} The answer is: \_\_\_\_ \\ \midrule

\textit{\ding{184}~Manual few-shot CoT prompting} \\
\vspace{-1mm}
\textbf{Q:} For the sentence pair "Who is winning the presidential election, Trump or Clinton?" and "Who's winning the election, Trump or Clinton?", do these two sentences have the same semantics?  \\
\textbf{A:} First, identify the key differences between the two sentences. Second, consider the impact of the difference in wording. Third, consider the overall meaning of the two sentences. Both sentences ask the same question about the current status of the election and the relative positions of the candidates. Therefore, given that the two sentences ask the same question and convey the same general meaning, despite the difference in wording and context, we can conclude that they have the same semantics. The answer (yes or no) is: yes. \\
\vspace{-1mm}
\textbf{Q:} For the sentence pair "How do I know if I really want to become a doctor?" and "How do I know if I want to be a doctor?", do these two sentences have the same semantics?  \\
\textbf{A:} First, identify the key differences between the two sentences. Second, consider the impact of the difference in wording. Third, consider the overall meaning of the two sentences. Both sentences ask how one can determine whether they want to become a doctor. However, the inclusion of "really" in the first sentence may imply a deeper level of introspection or a more significant decision. Therefore, given the difference in emphasis and potential implications, we can conclude that these two sentences do not have the same semantics. The answer (yes or no) is: no. \\ \hdashline
\vspace{-1mm}
\textbf{Q:} For the sentence pair "\textit{What are your favorite photos of yourself?}" and "\textit{Which is your favourite picture of yourself?}", do these two sentences have the same semantics? \\
\textbf{A:} The answer (yes or no) is: \_\_\_\_ \\ 

\bottomrule
    \end{tabular}
    \label{tab_qqp}
\end{table*}
\endgroup

\begingroup
\begin{table*}[htb]
    \centering
    \caption{
    Examples of standard few-shot prompting, zero-shot CoT prompting and manual few-shot CoT prompting produced by ChatGPT on the MNLI task. 
    }
    \vspace{2.8mm}
    \begin{tabular}{p{0.96\linewidth}}
    \toprule
\textit{\ding{182}~Standard few-shot prompting} \\
\vspace{-1mm}
\textbf{Q:} Given the sentence "well how do you think women's roles have changed" determine if the following statement is entailed or contradicted or neutral: "How have the roles of women changed?" \\
\textbf{A:} The answer (entailed or contradicted or neutral) is: entailed. \\
\vspace{-1mm}
\textbf{Q:} Given the sentence "Should further revelations warrant impeachment, Congress may have lost the necessary credibility." determine if the following statement is entailed or contradicted or neutral: "Further news will never do anything to stifle the notion of Congress credibility." \\
\textbf{A:} The answer (entailed or contradicted or neutral) is: contradicted. \\ \hdashline
\vspace{-1mm}
\textbf{Q:} Given the sentence "\textit{Others watched them with cold eyes and expressionless faces.}" determine if the following statement is entailed or contradicted or neutral: "\textit{Some people who were not emotive were watching.}" \\
\textbf{A:} The answer (entailed or contradicted or neutral) is: \_\_\_\_ \\ \midrule

\textit{\ding{183}~Zero-shot CoT prompting} \\
\vspace{-1mm}
\textbf{Q:} Given the sentence "\textit{Others watched them with cold eyes and expressionless faces.}" determine if the following statement is entailed or contradicted or neutral: "\textit{Some people who were not emotive were watching.}" \\
\textbf{A:} \textbf{\textit{Answer (entailed or contradicted or neutral) the question step by step.}} The answer is: \_\_\_\_ \\ \midrule

\textit{\ding{184}~Manual few-shot CoT prompting} \\
\vspace{-1mm}
\textbf{Q:} Given the sentence "well how do you think women's roles have changed" determine if the following statement is entailed or contradicted or neutral: "How have the roles of women changed?"  \\
\textbf{A:} First, identify the main proposition or claim of the original sentence. Second, identify the question being asked in the statement. Third, compare the main proposition or claim of the original sentence with the question being asked in the statement. The main proposition or claim of the original sentence and the question being asked in the statement are similar in meaning and are asking about the same topic. Lastly, determine the relationship between the original sentence and the statement. The statement is asking a question that is similar in meaning to the main proposition or claim of the original sentence. Therefore, we can conclude that the statement is entailed by the original sentence. The answer (entailed or contradicted or neutral) is: entailed. \\
\vspace{-1mm}
\textbf{Q:} Given the sentence "Should further revelations warrant impeachment, Congress may have lost the necessary credibility." determine if the following statement is entailed or contradicted or neutral: "Further news will never do anything to stifle the notion of Congress credibility."  \\
\textbf{A:} First, identify the main proposition or claim of the original sentence. Second, identify the question being asked in the statement. Third, compare the main proposition or claim of the original sentence with the question being asked in the statement. The main proposition of the original sentence and the claim in the statement are contradictory because the original sentence implies that further revelations could harm Congress's credibility, while the statement suggests that they could not. Lastly, determine the relationship between the original sentence and the statement. The statement contradicts the main proposition or claim of the original sentence. Therefore, we can conclude that the statement is contradicted by the original sentence. The answer (entailed or contradicted or neutral) is: contradicted. \\ \hdashline
\vspace{-1mm}
\textbf{Q:} Given the sentence "\textit{Others watched them with cold eyes and expressionless faces.}" determine if the following statement is entailed or contradicted or neutral: "\textit{Some people who were not emotive were watching.}" \\
\textbf{A:} The answer (entailed or contradicted or neutral) is: \_\_\_\_ \\
\bottomrule
    \end{tabular}
    \label{tab_mnli}
\end{table*}
\endgroup

\begingroup
\begin{table*}[htb]
    \centering
    \caption{
    Examples of standard few-shot prompting, zero-shot CoT prompting and manual few-shot CoT prompting produced by ChatGPT on the QNLI task. 
    }
    \vspace{2.8mm}
    \begin{tabular}{p{0.96\linewidth}}
    \toprule
\textit{\ding{182}~Standard few-shot prompting} \\
\vspace{-1mm}
\textbf{Q:} Given the question "The "To Secure These Rights" findings wanted desegregation of which institution?" determine if the following sentence contains the corresponding answer: "These Rights also called for desegregation of the Armed Forces." \\
\textbf{A:} The answer (yes or no) is: yes. \\
\vspace{-1mm}
\textbf{Q:} Given the question "Where is the Bay of Pasaia located?" determine if the following sentence contains the corresponding answer: "A series of expeditions to St. John's by Portuguese from the Azores took place in the early 16th century, and by 1540 French, Spanish and Portuguese ships crossed the Atlantic annually to fish the waters off the Avalon Peninsula." \\
\textbf{A:} The answer (yes or no) is: no. \\ \hdashline
\vspace{-1mm}
\textbf{Q:} Given the question "\textit{Where is English spoken the most?}" determine if the following sentence contains the corresponding answer: "\textit{English is widely spoken in commerce, schooling and government.}" \\
\textbf{A:} The answer (yes or no) is: \_\_\_\_ \\ \midrule

\textit{\ding{183}~Zero-shot CoT prompting} \\
\vspace{-1mm}
\textbf{Q:} Given the question "\textit{Where is English spoken the most?}" determine if the following sentence contains the corresponding answer: "\textit{English is widely spoken in commerce, schooling and government.}" \\
\textbf{A:} \textbf{\textit{Answer (yes or no) the question step by step.}} The answer is: \_\_\_\_ \\ \midrule

\textit{\ding{184}~Manual few-shot CoT prompting} \\
\vspace{-1mm}
\textbf{Q:} Given the question "How large was the fleet that attacked the Bithynian coast along with the Pecheneg allies?" determine if the following sentence contains the corresponding answer: "A navy of 10,000 vessels, including Pecheneg allies, landed on the Bithynian coast and devastated the Asiatic shore of the Bosphorus."  \\
\textbf{A:} First, identify the key information in the question: "How large was the fleet that attacked the Bithynian coast along with the Pecheneg allies?" Second, look for the corresponding information in the given sentence: "A navy of 10,000 vessels, including Pecheneg allies, landed on the Bithynian coast and devastated the Asiatic shore of the Bosphorus." Third, compare the key information in the question to the corresponding information in the sentence to determine if the sentence contains the answer. The fleet that attacked the Bithynian coast along with the Pecheneg allies was a navy of 10,000 vessels, according to the sentence. Therefore, we can conclude that the sentence contains the answer. The answer (yes or no) is: yes. \\
\vspace{-1mm}
\textbf{Q:} Given the question "Heading east from Southampton, what city is connected by rail?" determine if the following sentence contains the corresponding answer: "The route to London was opened in 1840 by what was to become the London and South Western Railway Company."  \\
\textbf{A:} First, identify the key information in the question: "Heading east from Southampton, what city is connected by rail?" Second, look for the corresponding information in the given sentence: "The route to London was opened in 1840 by what was to become the London and South Western Railway Company." Third, compare the key information in the question to the corresponding information in the sentence to determine if the sentence contains the answer. The sentence mentions that the route to London was opened by the London and South Western Railway Company, but it does not mention if there are any other cities connected by rail from Southampton. Therefore, we can conclude that the sentence does not contain the corresponding answer to the question. The answer (yes or no) is: no. \\ \hdashline
\vspace{-1mm}
\textbf{Q:} Given the question "\textit{Where is English spoken the most?}" determine if the following sentence contains the corresponding answer: "\textit{English is widely spoken in commerce, schooling and government.}" \\
\textbf{A:} The answer (yes or no) is: \_\_\_\_ \\

\bottomrule
    \end{tabular}
    \label{tab_qnli}
\end{table*}
\endgroup

\begingroup
\begin{table*}[htb]
    \centering
    \caption{
    Examples of standard few-shot prompting, zero-shot CoT prompting and manual few-shot CoT prompting produced by ChatGPT on the RTE task. 
    }
    \vspace{2.8mm}
    \begin{tabular}{p{0.96\linewidth}}
    \toprule
\textit{\ding{182}~Standard few-shot prompting} \\
\vspace{-1mm}
\textbf{Q:} Given the sentence "Harvey Weinstein, the co-chairman of Miramax, who was instrumental in popularizing both independent and foreign films with broad audiences, agrees." determine if the following statement is entailed: "Harvey Weinstein is the co-chairman of Miramax." \\
\textbf{A:} The answer (yes or no) is: yes. \\
\vspace{-1mm}
\textbf{Q:} Given the sentence "INS predicts that the smuggling will continue to increase and that alien smuggling organizations will become more sophisticated, organized, and complex." determine if the following statement is entailed: "Steps are being taken to stop the smuggling of aliens." \\
\textbf{A:} The answer (yes or no) is: no. \\ \hdashline
\vspace{-1mm}
\textbf{Q:} Given the sentence "\textit{Hepburn's family will receive the proceeds from the sale.}" determine if the following statement is entailed: "\textit{Proceeds go to Hepburn's family.}" \\
\textbf{A:} The answer (yes or no) is: \_\_\_\_ \\ \midrule

\textit{\ding{183}~Zero-shot CoT prompting} \\
\vspace{-1mm}
\textbf{Q:} Given the sentence "\textit{Hepburn's family will receive the proceeds from the sale.}" determine if the following statement is entailed: "\textit{Proceeds go to Hepburn's family.}" \\
\textbf{A:} \textbf{\textit{Answer (yes or no) the question step by step.}} The answer is: \_\_\_\_ \\ \midrule

\textit{\ding{184}~Manual few-shot CoT prompting} \\
\vspace{-1mm}
\textbf{Q:} Given the sentence "Anna Politkovskaya was found shot dead on Saturday in a lift at her block of flats in the Russian capital, Moscow." determine if the following statement is entailed: "Anna Politkovskaya was murdered."  \\
\textbf{A:} First, is there any mention of Anna Politkovskaya's death in the sentence? Yes, the sentence mentions that Anna Politkovskaya was found shot dead. Second: does the sentence suggest how Anna Politkovskaya died? Yes, the sentence suggests that Anna Politkovskaya died from being shot. Third, does the sentence use any words that indicate that Anna Politkovskaya's death was intentional or deliberate? Yes, the sentence uses the word "shot" which suggests that Anna Politkovskaya's death was intentional. Lastly, based on the information provided in the sentence, can we conclude that Anna Politkovskaya was murdered? Yes, we can conclude that Anna Politkovskaya was murdered because she was found shot dead, which suggests that someone intentionally and unlawfully caused her death. Therefore, the answer (yes or no) is: yes. \\
\vspace{-1mm}
\textbf{Q:} Given the sentence "Oscar-winning director Franco Zeffirelli has been awarded an honorary knighthood for his "valuable services to British performing arts"." determine if the following statement is entailed: "Italian director is awarded an honorary Oscar."  \\
\textbf{A:} First, is there any mention of an Oscar in the sentence? No, there is no mention of an Oscar in the sentence. Second, does the sentence suggest that Franco Zeffirelli received any award related to film or cinema? No, the sentence mentions that Franco Zeffirelli was awarded an honorary knighthood for his services to the British performing arts, but there is no indication that he received an honorary Oscar. Therefore, based on the information provided in the sentence, we cannot conclude that Franco Zeffirelli was awarded an honorary Oscar. Therefore, the answer (yes or no) is no.  \\ \hdashline
\vspace{-1mm}
\textbf{Q:} Given the sentence "\textit{Hepburn's family will receive the proceeds from the sale.}" determine if the following statement is entailed: "\textit{Proceeds go to Hepburn's family.}" \\
\textbf{A:} The answer (yes or no) is: \_\_\_\_ \\

\bottomrule
    \end{tabular}
    \label{tab_rte}
\end{table*}
\endgroup

\end{document}